\documentclass[a4paper, 10pt, journal]{ieeeconf}   
\usepackage[utf8]{inputenc}
\usepackage{dtk-logos} 

\usepackage{caption}
\usepackage{url}
\usepackage{graphicx, subfig}
\usepackage{booktabs}
\usepackage{threeparttable}
\usepackage{footnote}
\usepackage{stfloats}
\title{\LARGE \bf
MIDI-Sandwich2: RNN-based Hierarchical Multi-modal Fusion Generation VAE networks for multi-track symbolic music generation
\thanks{Supported by the National Key Research and Development Program of China under Grant 2018YFB1003601.}}

\author{Xia Liang$^{1}$ and Junmin Wu$^{2}$ and Jing Cao\\
	University of Science and Technology of China\\
	Hefei, Anhui, China \\
	$^{1}$sa517190@mail.ustc.edu.cn
	$^{2}$jmwu@ustc.edu.cn
}

\begin{document}

\maketitle


\begin{abstract}
Currently, almost all the multi-track music generation models use the Convolutional Neural Network (CNN) to build the generative model, while the Recurrent Neural Network (RNN) based models can not be applied in this task. In view of the above problem, this paper proposes a RNN-based Hierarchical Multi-modal Fusion Generation Variational Autoencoder (VAE) network, MIDI-Sandwich2, for multi-track symbolic music generation. Inspired by VQ-VAE2\cite{razavi2019generating}, MIDI-Sandwich2 expands the dimension of the original hierarchical model by using multiple independent Binary Variational Autoencoder (BVAE) models without sharing weights to process the information of each track. Then, with multi-modal fusion technology, the upper layer named Multi-modal Fusion Generation VAE (MFG-VAE) combines the latent space vectors generated by the respective tracks, and uses the decoder to perform the ascending dimension reconstruction to simulate the inverse operation of multi-modal fusion, multi-modal generation, so as to realize the RNN-based multi-track symbolic music generation. For the multi-track  format \emph{pianoroll}, we also improve the output binarization method of MuseGAN\cite{dong2018musegan}\cite{dong2018convolutional}, which solves the problem that the refinement step of the original scheme is difficult to differentiate and the gradient is hard to descent, making the generated song more expressive. The model is validated on the Lakh Pianoroll Dataset (LPD) multi-track dataset. Compared to the MuseGAN, MIDI-Sandwich2 can not only generate harmonious multi-track music, the generation quality is also close to the state of the art level. At the same time, by using the VAE to restore songs, the semi-generated songs reproduced by the MIDI-Sandwich2 are more beautiful than the pure autogeneration music generated by MuseGAN. Both the code and the audition audio samples are open source on \url{https://github.com/LiangHsia/MIDI-S2}.

\end{abstract}

\begin{keywords}
Deep learning, Multi-track symbolic music generation, Multi-modal fusion generation, Hierarchical VAE architecture.
\end{keywords}

\section{Introduction}

With the development of deep learning in the field of symbolic music generation, the generated products vary from single-track music to polyphonic music to multi-track music\cite{briot2017deep}. However, according to the characteristics of multi-track symbolic music’s data representation, the multi-track generation models generally regard the \emph{pianoroll} format of music as an image, and regard different tracks as different channels, and then they process music using the Convolutional Neural Network\cite{krizhevsky2012imagenet} (CNN) model. For example, MIDI-VAE\cite{brunner2018midi} and Cycle-GAN\cite{brunner2018symbolic} use CNN to build a model that can perform style transformation on multi-track music. Xiaoice band\cite{zhu2018xiaoice} has achieved good results on the multi-track music generation problem. It generates different tracks in a certain order, and uses the previous generated track as the constraint of the latter track to achieve harmony between the tracks. However, this method cannot generate multiple tracks at the same time, and the restriction between tracks is difficult to unify.

MuseGAN\cite{dong2018musegan} uses CNN-based Generative Adversarial Networks\cite{goodfellow2014generative} (GAN) to achieve multi-track music generation. It has also built its own multi-track dataset, Lakh Pianoroll Dataset, which sets the music sequence to the data format of the \emph{pianoroll}. The upgraded version of MuseGAN, BMuseGAN\cite{dong2018convolutional}, binarizes the output format to better fit the data format and thus optimizing the model generation result. However, the binarization method proposed by BMuseGan has difficulty in differentiating the key steps of binarization, which makes the convergence of the model harder.

The Variational Auto-encoder\cite{doersch2016tutorial} (VAE) is a generator that performs as well as the GAN. The Hierarchical Variational Autoencoder model, VQ-VAE2\cite{razavi2019generating}, proposed by \emph{DeepMind} recently, can generates diverse high-definition images. At the same time, MIDI-Sandwich$\footnote{Midi-Sandwich: https://arxiv.org/abs/1907.01607}$ also proposes a similar hierarchical VAE-GAN model to generate a longer, more structured single track music. By sharing the weight of the underlying VAE, it uses the underlying VAE to process local information. Then the upper layer uses Global VAE to explore the structural information between local spaces by analyzing the latent space vector generated by the underlying layer. 

Compared to single-track music, the data representation of multi-track symbolic music has one more dimension named track. As a result, almost all the multi-track music generation models use the Convolutional Neural Network (CNN) to build the generative model, while the Recurrent Neural Network (RNN) based models can not be applied in this task.
As for the Recurrent Neural Network\cite{mikolov2010recurrent} (RNN) based symbolic music generation model, although there are a series of models represented by Google Magenta\cite{casella2001magenta}, the multi-track music generation problem is not solved perfectly. 

In view of this problem, this paper proposes a RNN-based hierarchical multi-modal fusion generation VAE networks, MIDI-Sandwich2. Inspired by VQ-VAE2 and Midi-Sandwich, MIDI-Sandwich2 uses a hierarchical architecture. The difference is that its underlying model uses an independent, unshared weights VAE named Bianary VAE (BVAE) to handle different multi-track music information. The upper model model, Multi-modal Fusion Generation VAE (MFG-VAE), uses a encoder as a multi-modal fusion module to project the latent vector of different underlying spaces into the same low-dimensional space. The decoder is seen as the inverse operation of multi-modal fusion to reconstruct (generate) multi-modal generation products. MIDI-Sandwich2 uses a multi-modal simultaneous generation method to combine individual single RNN modules to collaboratively generate harmonious multi-track music. At the same time, for the multi-track \emph{pianoroll} format, this paper makes an improvement on BMuseGAN's refine method. It converts the binary output problem into a multi-label classification problem\cite{tsoumakas2007multi}, which solves the problem that the refinement step of the original scheme is difficult to differentiate and the gradient is hard to descent.

The experimental results show that our model can effectively generate multi-track music. Compared with MuseGAN, the music directly generated by MIDI-Sandwich2 is close to the level of MuseGAN, while the songs adapted by VAE's reductive ability are of better quality than that of MuseGAN, which is of great significance for the development of RNN in multi-track music.   

This paper makes the following contributions:
 
1. It proposes a hierarchical Multi-modal Fusion Generation VAE framework. The model extends the processing dimension of VQ-VAE2 and MIDI-Sandwich’s hierarchical architecture. By using the independent underlying VAE to process different modal information, and the VAE's encoder-decoder architecture, it realizes inverse operation of multi-modal fusion, multi-modal generation.

2. It uses the multi-modal fusion generation VAE framework to tackle multitrack symbolic music generation problem. By using the RNN-based networks, it breaks the boundary that RNN-based sequence model can not generate multi-track music.

3. For symbolic multi-track music format \emph{pianoroll}, MIDI-Sandwich2 makes improvements to the BMuseGAN’s binarization strategy. By converting the target task into a multi-label classification task, it achieves the binarization of the network output. As a result, it solves the problem that the refinement step of the original scheme is difficult to differentiate and the gradient is hard to descent.

\section{Related Work}

\subsection{MuseGAN, BMuseGAN}
Musegan\cite{dong2018musegan}\cite{raffel2016learning} has made a significant contribution to the generation of multi-track symbolic music. It built its own dataset, Lakh Pianoroll Dataset (LPD). The music format in the dataset is \emph{pianoroll}. \emph{Pianoroll} is a multi-dimensional matrix of (timestep, pitch, track). If a sound is pressed, the value of the sound is set to 1, while the other values are -1. MuseGAN has also defined a series of evaluation methods to measure the generated music objectively. It has also proposed a CNN-based multi-track generation model using WGAN-GP\cite{gulrajani2017improved} network. 

The upgraded version of Musegan, BMuseGAN, proposes the concept of binarized output to process the LPD data format. It proves that during the training phase, instead of real values, fixing the value of the generated pianroll at -1 and 1 can effectively improve the quality of generated music. BMuseGAN proposes a refine networks, which can perform binarization of the generated network output. However, the refinement step of BMuseGAN is difficult to differentiate and the gradient is hard to descent.

Binary Neurons (BNs) are neurons that output binary-valued predictions. The BMuseGAN consider two types of BNs: deterministic binary neurons (DBNs) and stochasticbinary neurons (SBNs). DBNs act like neurons with hard \emph{thresholding} functions as their activation functions. The BMuseGAN defines the output of a DBN for a real-valued input $x$ as:
\begin{equation}
DBS(x)=u(\sigma(x)-0.5),
\end{equation}
where $u$ is the unit step function and $\sigma$ is the logistic sigmoid function.  SBNs, in contrast, binarize aninputxaccording to a probability, defined as:
\begin{equation}
SBN(x)=u(\sigma(x)-v), v\sim U[0,1],
\end{equation}
where $U[0,1]$ denotes a uniform distribution.

Computing the exact gradients for either DBNs or SBNs, however, is intractable. For SBNs, it requires the computation of the average loss over all possible binary samplings of all the SBNs, which is exponential in the total numberof SBNs. For DBNs, the threshold function is non-differentiable.
\subsection{ Multi-modal Fusion}
The Multi-modal fusion\cite{atrey2010multimodal} problem is to extract relevant effective information in data features of multiple different modalities, thus improving the final decision-making ability. Multi-modal fusion is divided into three categories under different fusion time: feature level fusion, decision level fusion, and hybrid fusion. 

This paper mainly focuses on the feature level. Feature layer fusion is to transform multi-dimensional features into single-dimensional features, and then deal with the merged features. The detailed methods are subject to the problem. Generally, there are two approaches: using the feature engineering to directly calculate the projection matrix, or using a deep learning neural network to optimize a low-dimensional space.
\subsection{VAE, Vq-vae2, MIDI-Sandwich}
The normally VAE\cite{doersch2016tutorial} consists of an encoder and a decoder. The VQ-VAE2 proposed by \emph{DeepMind} recently has surpassed the latest GAN generator BIG GAN\cite{brock2018large} in image generation. VQ-VAE2 is a hierarchical network architecture. The bottom layer uses multiple CVAEs to process local information. The CVAE shares weights with each other. The upper layer VAE performs global control on the generated data by analyzing the latent vector from the underlying CVAE. The network has been proven to be able to produce larger, higher resolution, sharper and finer images. 

MIDI-Sandwich is also presented at the same time. It uses hierarchical VAE-GANs to generate longer, more structured single-track music sequences. Different from the above two methods, the underlying VAE in this paper no longer share weights and deal with the same modality single-dimension. According to different problems, MIDI-Sandwich2 can handle information of different dimensions through multiple independent VAEs, thus achieving feature presentation, fusion and generation tasks of multi-modal.

\section{MIDI-Sandwich2}
This paper proposes a RNN-based Hierarchical Multi-modal Fusion Generation VAE networks for the generation of multi-track symbolic music. The model is divided into two layers: the bottom layer named Binary VAE (BVAE) and the upper layer named Multi-modal Fusion Generation VAE (MFG-VAE).

\begin{figure}
	\centering
	\includegraphics[width=3.0in]{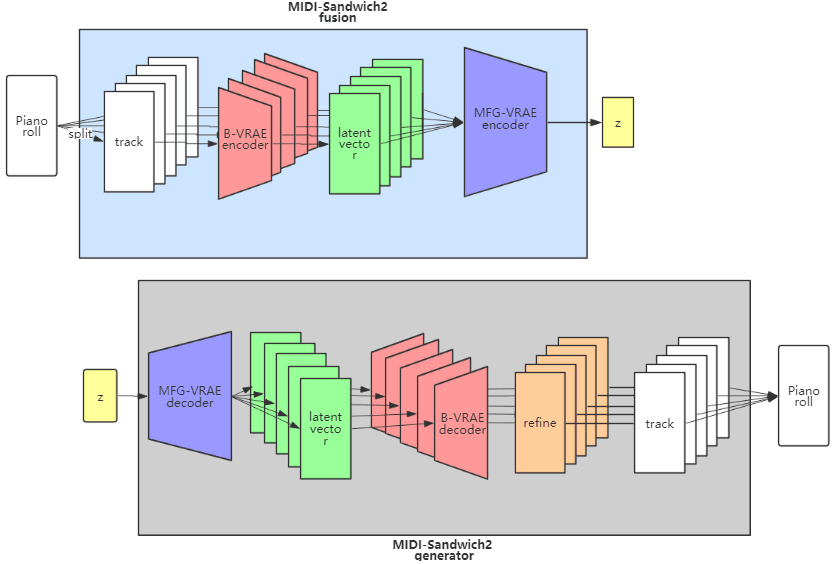}
	\caption{The MIDI-Sandwich2 architecture}
	\label{fig:side:a}
\end{figure}

As shown in Figure \ref{fig:side:a}, the hierarchical VAE network firstly uses multiple independent BVAEs' encoder layers to extract feature information of instruments in different tracks. Then, MFG-VAE fuses and generates features of these different modals. Finally, by using the reconstructed multi-modal feature, the BVAE's decoder generates multi-track music.

Midi-Sandwich2 has two ways to generate music: 1) The input is \emph{pianoroll} and MIDI-Sandwich2 reproduces it using the full VAE architecture; 2) The input is a subject to Gaussian random variable and MIDI-Sandwich2 decodes it into a \emph{pianoroll}.

The training process is divided into two steps: stage 1, training B-VAE, and stage 2, training MIDI-Sandwich2 composed of BVAE and MFG-VAE. 

\subsection{Binary VAE} 
BVAE uses Variational recurrent auto-encoders (VRAE)\cite{fabius2014variational}, which is based on Long Short-Term Memory (LSTM), to process each track and learn the playing modes of different instruments, such as drums, guitars and basses. Moreover, like BMuseGAN, it uses a refine network to generate binarized music.

The Binary Refine module proposed in this paper converts the [-1, 1] binarization task into a multi-sample classification task. The specific steps are as follows. Firstly, it converts the [-1, 1] binary value into [0, 1], so that the single-track pitch vector at each timestep in the data set can also be regarded as a multi-label classification tag. At the same time, a sigmoid layer is added at the end of the binary refine module network. It performs sigmoid calculation for data in the network’s last dimension, so that the output value range of the network is reduced to between 0 and 1. Finally, we use the \emph{binary crossentropy} of the multi-label classification task as the loss function. 

This series of operations converts the output binarization task into a multi-label classification task. Compared with BmuseGAN, the gradient descent and back propagation can be performed normally.

\begin{figure}
	\centering
	\includegraphics[width=3.0in]{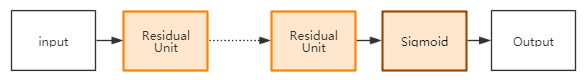}
	\caption{The refine network. The tensor size remains the same throughout the network.}
	\label{fig:side:bb}
\end{figure}

As shown in Figure \ref{fig:side:bb}, the refine model first considers the output generated by VRAE as a picture, and extracts features through several Risidual Unit layers, and adds a sigmoid layer to the last layer of the network. In addition, the training set of the refined model is slightly different from the original training set. We need to convert the [-1, 1] binary value in the original dataset into [0, 1] binary value. Then they are used as the $y$ tag for multi-label classification. The $y$ tag conversion formula from the original dataset to multi-label classification is as follows:

\begin{equation}
y=(y_{[-1,1]}+1)/2, 
\end{equation}
where $y_{[-1,1]}$ represents the original data set and $y$ represents the data set required by the converted refine model.
The following is the loss function (\emph{binary crossentropy}) to refine multi-label classification.

\begin{equation}
L=-\frac{1}{N}\sum_{i=1}^N [{y_i} \log(\hat{y}_i)+(1-y_i) \log(1-\hat{y}_i)],
\end{equation}
where $N$ is the label dimension and $\hat{y}_i$ is the predicted value of the model.
\subsection{The Multi-modal Fusion Generation VAE}
MIDI-Sandwich2 proposes three multi-modal fusion generation strategies, and finally uses the one with the best result as the actual generation strategy.

1. The model directly concatenates multiple features to achieve multi-modal fusion, then uses Multi-Layer Perception (MLP) to perform dimensional reduction sampling and up-dimensional decoding. This method is widely used in deep learning\cite{ngiam2011multimodal}.

2. The model directly calculates the vector obtained by projecting multiple features into the low-dimensional space, and then uses the MLP to perform dimensional reduction sampling and up-dimensional decoding. 

3. The model combines the multi-modal feature vectors into a two-dimensional matrix of (track, feature vector), and then use VRAE with the similar architecture as the underlying BVAE to achieve the fusion and generation of the latent vector feature matrix.

As shown in Figure 3, different fusion generation strategies are essentially VAE's encoder and decoder operations. When using the VAE method, VAE will first reduce the dimension, then learn the mean value $\mu$ and the variance $\sigma$ of the latent vector of MFG-VAE, and add an error matrix $\epsilon$ to improve the robustness of the random generation. The final sampled descending dimension latent vector subject to Gaussian distribution. The sampling formula is as follows:

\begin{equation}
z = \mu(X)+\sigma^\frac{1}{2}(X)*\epsilon, \epsilon \sim N(0,1).
\end{equation}

Therefore, the theory put forward in this paper is to combine multi-modal fusion with VAE's encoder. By using multi-modal fusion to learn the potential commonality between the modals, we sample the commonality's latent vector representation subject to Gaussian distribution. Finally, we use the decoder to restore the latent vector matrix, thus improving decoder's ability to generate multi-modal artifacts.

the Multi-modal Fusion Generation VAE loss fuction is defined as follows: 
\begin{equation}
L=E_{X\sim \bar{p} (X)}[E_{z\sim p(Z|X)}[-\ln q(X|Z)]+KL(p(Z|X)||q(Z))],
\end{equation}
where the first and second terms are the reconstruction loss and a prior regularization term that is the Kullback-Leibler (KL) divergence, respectively.

\begin{figure}
	\centering
	\includegraphics[width=3.0in]{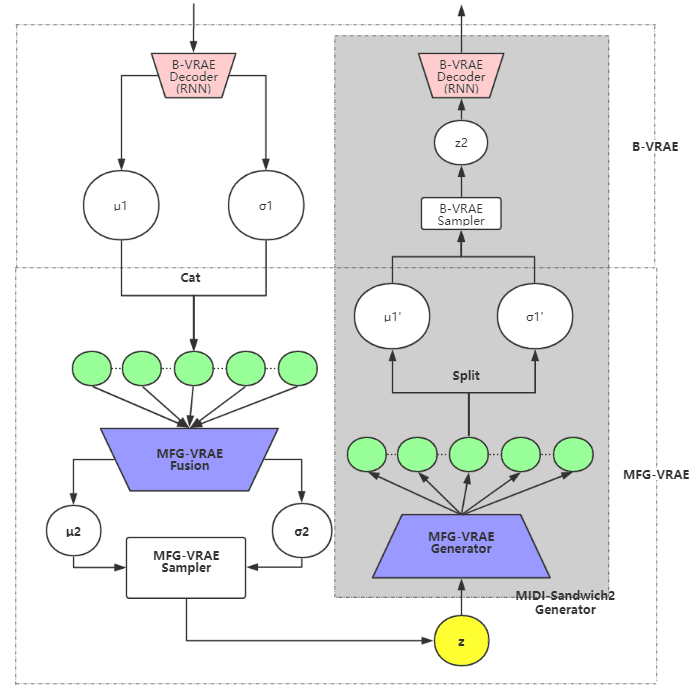}
	\caption{Multi-modal Fusion Generation VAE}
	\label{fig:side:c}
\end{figure}

\subsection{The combination strategy of BVAE and MFG-VAE}

In MIDI-Sandwich2, MFG-VAE is in the upper layer, and BVAE is in the bottom layer. There are two points to note when combining them.

1 We should train BVAE firstly in the first stage. After the training of BVAE is finished, the second stage will freeze BVAE and then start the training of MFG-VAE.

2 BVAE is best generated when the input $x$ of decoder subject to Gaussian distribution. Therefore, the latent vector analyzed by the upper layer MFG-VAE is actually the mean vector and the variance vector before the BVAE sample. The BVAE sample is placed behind the step that MFG-VAE reconstructs the multi-dimensional latent vector. There are two reasons for doing so. Firstly, the sampling of the BVAE introduces a random error, which affects MFG-VAE’s analysis of the latent vector. Secondly, by performing sampling, we can get the vector that subject to the Gaussian distribution, which can be used as the input of the BVAE decoder to improve the quality of the generation.

\section{Experiment}
\subsection{Experiment Details}
This paper uses the LPD multi-track dataset introduced in Section 2, which is almost the best multi-track symbolic music dataset currently. The dataset has five tracks (bass, drums, guitar, strings and piano) fixed for each piece of music, and each piece of music is divided into four consecutive bars. To compress the data set size, the actual data storage unit is binary, not real.

The paper does not do much hyperparameter tuning work. The BVAE's latent vector dimension is set to 64, and the MFG-VAE's latent vector dimension is set to 128. The \emph{batchsize} for both training phases is set to 32. The intermediate dim of VRAE is set to 128. The \emph{standard epsilon} of BVAE and MFS-VAE are both set to 0.01. The size of convolutional kernal,  which is part of residual unit used in the refiner network, is (1, 3,  12).

The experiment is performed on the \emph{NVIDIA 16G P100} and the complete experiment can be finished in one day. 

\section{Result}
We employ to MuseGAN's Objective Evaluation to evaluate the quality of the music generated by the MIDI-Sandwich2. To verfy our work, we analyze the pictures of \emph{pianoroll} and the loss process of all stages. Finally, we use random voting to reflect some audience's evaluation of our model and MuseGAN.
\subsection{Objective Evaluation} 
We employ the MuseGAN evaluation method Objective Metrics for evaluation. We evaluate the real experimental set (dataset), generator of MIDI-Sandwich2 without refine networks, generator of MIDI-Sandwich2 and reducer of MIDI-Sandwich2 . The result is shown in Table \ref{tab:aStrangeTable1}.

\begin{table*}[!htbp]
\	
	\centering
	\caption{Objective Metrics Evaluation(SR: scale ratio(in \%), UPC: number of used pitch classes per bar (from 0 to 12), QN: ratio of "qualified" notes, DP, or drum pattern: ratio of notes in 8- or 16-beat pat-terns, common ones for Rock songs in 4/4 time (in \%). \textbf{Values closer to the first row are better.}}\label{tab:aStrangeTable1}
	\begin{tabular}{|c|ccccc|cccc|c|}
		\hline
		\multicolumn{1}{|c|}{}
		&\multicolumn{5}{c|}{scale ratio (SR; \%)}
		&\multicolumn{4}{c|}{used pitch classes (UPC)}
		&\multicolumn{1}{c|}{DP (\%)} \\
		\multicolumn{1}{|c|}{} & B&D&G&P&S&B&G&P&S&D\\
		\hline \hline
		Dataset &60.4 & 72.1 & 71.7 & 46.1 &71.4 & 15 & 75 & 41 & 82 &94.4\\
		\hline
		MSG1 &97.6 & 73.7 & 70.5 & 35.8 &80.0 & 4 & 106 & \textbf{47} & 66 &98.5\\
		MSG2 &53.0 & 72.6 & 80.2 & 43.5 &\textbf{73.7} & \textbf{16} & 88 & 37 & 70 &\textbf{97.4}\\
		MSR  &\textbf{55.9} & \textbf{71.6} & \textbf{72.5} & \textbf{50.3} &77.3 & \textbf{14} & \textbf{72} & 48 & \textbf{78} &97.6\\ 
		\hline
	\end{tabular}
    \footnotesize{\\ \textbf{MSG1}: from MIDI-Sandwich2 generator (without Refine). \textbf{MSG2}: from MIDI-Sandwich2 generator.\\ \textbf{MSR}: from MIDI-Sandwich2 reducer.}\\
	
\end{table*}

As can be seen from the Table \ref{tab:aStrangeTable1}, After adding refine networks, each index has been improved (closer to the Dataset), and the indexes reconstructed by MIDI-Sandwich2 reducer are closest to the original data.

\subsection{Training Process}

\begin{figure*}[!bp]
	\begin{minipage}[t]{0.5\linewidth}
		\centering
		\includegraphics[width=3.0in]{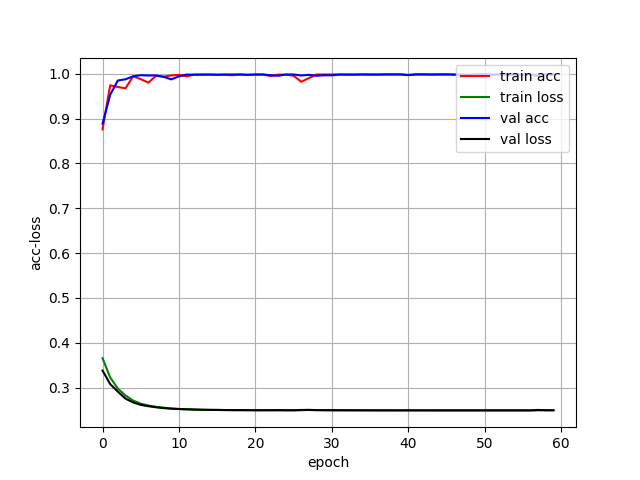}
		\caption{Training loss of refine network in notes.}
		\label{fig:side:bns}
	\end{minipage}%
	\begin{minipage}[t]{0.5\linewidth}
		\centering
		\includegraphics[width=3.0in]{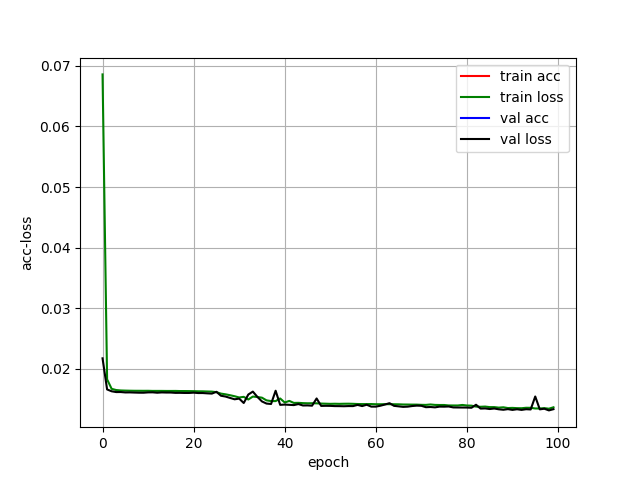}
		\caption{Training loss of BVAE networks in single track.}
		\label{fig:side:s1}
	\end{minipage}
	
\end{figure*}

\begin{figure*}[!bp]
	\begin{minipage}[t]{0.5\linewidth}
		\centering
		\includegraphics[width=3.0in]{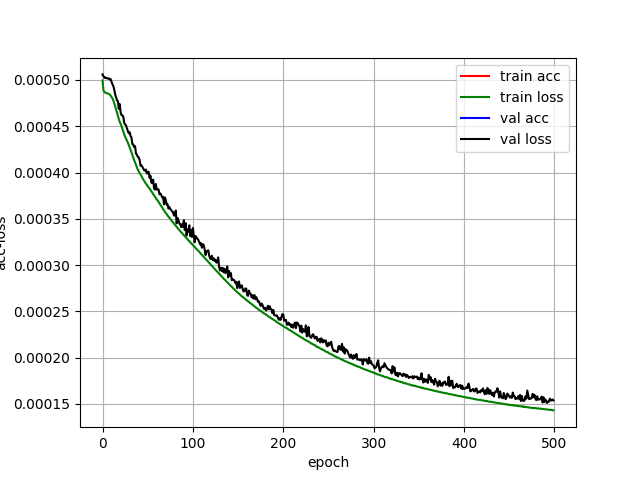}
		\caption{Training loss of strategy 3 of MFG-VAE networks.}
		\label{fig:side:s2}
	\end{minipage}%
	\begin{minipage}[t]{0.5\linewidth}
		\centering
		\includegraphics[width=3.0in]{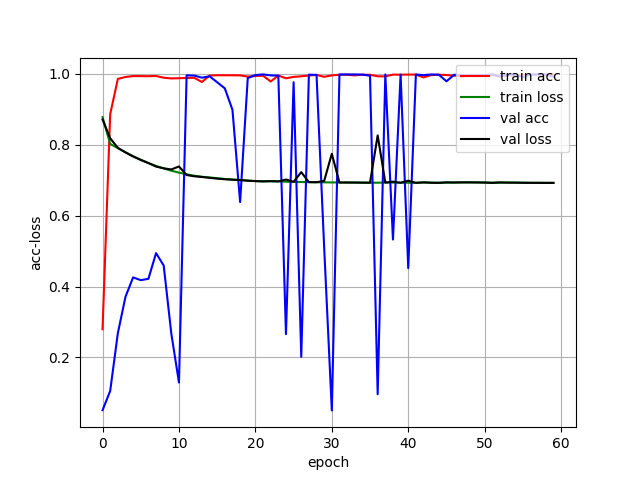}
		\caption{Training loss of refine network in bar.}
		\label{fig:side:bns2}
	\end{minipage}
\end{figure*}

As we can see in Figure \ref{fig:side:bns}, our proposed refine network has solved the problem that the original BmuseGAN cannot differentiate and gradient in. The refined network can be trained well. Figure \ref{fig:side:s1} shows that when training a BVAE for a single track, the model converges quickly and the loss is relatively stable. Figure \ref{fig:side:s2} shows that training loss of strategy 3 of MFG-VAE networks. Through the current experiment, the effect of strategy 3, using VRAE, is the best of the three strategy. the model converges slowly on the fusion generatation multi-track task, but the loss is even lower than that of Figure \ref{fig:side:s1}. 

Figure \ref{fig:side:bns2} is a refine network trained in bar. The acc of val is fluctuating. Part of the reason is that multi-label classification is not suitable for tasks with huge label dimensions. Therefore, our final refine network is selected to do multi-label classification in notes.
\begin{figure*}[!bp]
	\centering
	\includegraphics[width=5.0in]{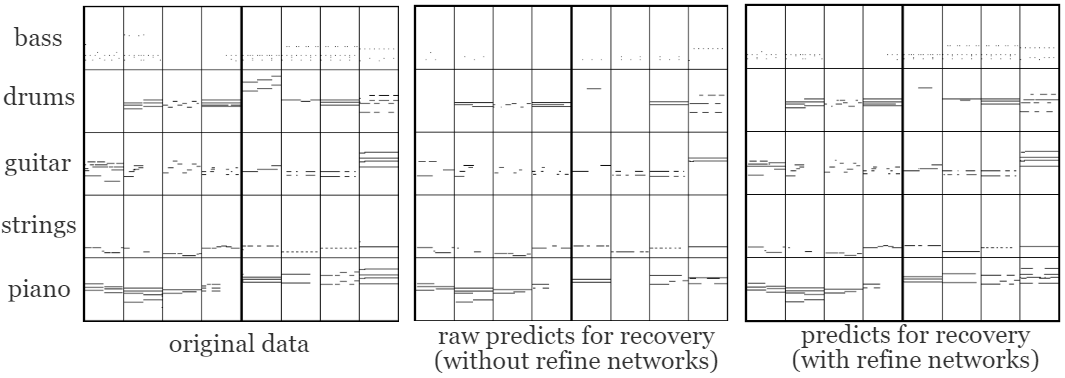}
	\caption{The influence of refine networks is shown on \emph{pianoroll}}
	\label{fig:side:bn}
\end{figure*}

\begin{figure*}[!bp]
	\centering
	\includegraphics[width=5.0in]{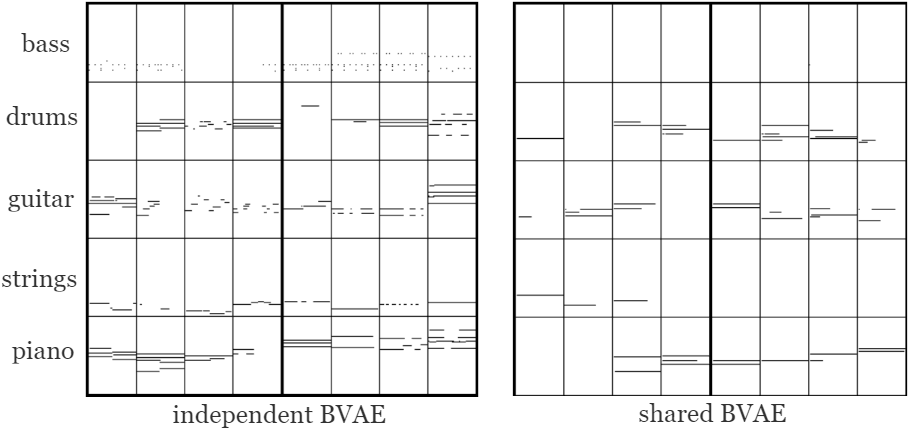}
	\caption{\emph{Pianoroll} of independent BVAE and shared weights BVAE}
	\label{fig:side:in}
\end{figure*}
As shown in Figure \ref{fig:side:bn}, there are the original data, the MIDI-Sandwich2 restore product without refinement and the complete MIDI-Sandwich2 restore product of the \emph{pianoroll} diagram, respectively. It is clear that the refine network is filled with more missing information and effective information, thus making the generated music more complete and rich. At the same time, although the overall structure of the reduction product and the original data are similar, they are slightly different. This phenomenon is equivalent to adaptation in the music field and is a generation phenomenon can never found in other generation networks such as GAN or LSTM.

Figure \ref{fig:side:in} are \emph{pianoroll} pictures using a shared BVAE as the underlying VAE and an independent BVAE as the underlying VAE, respectively. Obviously, compared to the shared BVAE, the independent BVAE can learn and generate the unique performance mode of each instrument through MFG-VAE. This shows that MIDI-Sandwich2 is effective in adding the underlying VAE dimension to the multi-track problem. At the same time, we also compared the music generated directly by the independent BVAE without fusion. The effect of multi-track harmony can not be directly reflected on the \emph{pianoroll}, so we put it in the next user evaluation part.

\subsection{User Evaluation}

\begin{table*}[!htbp]
	\centering
	\caption{User Evaluation (R: Rhythm, M: Melody, I: Integrity, St: Structure, H: Harmonious, MM: Melodic Motion.)}\label{tab:aStrangeTable3}
	\begin{tabular}{c|cccccc|c}
		\hline\hline
		Methods   & R& M& I& St &H & MM &Average\\
		\hline
		MuseGAN& \textbf{4.12}& \textbf{3.98} &3.12 &\textbf{3.57} &3.48 &2.55 &3.47\\
		MidiNet(GAN)& 3.89& 3.29 &2.68 &3.14 &2.96 &2.18 &3.02\\
		\hline
		 MIDI-Sandwich2(no fusion)&2.92& 2.43 &2.58 &2.13 &2.32 &2.46 &2.46\\
		 MIDI-Sandwich2(complete generation)& 3.82& 3.14 &3.38 &2.65 &3.27 & 2.12&3.05\\
		 MIDI-Sandwich2(complete restore)& \textbf{4.26}& 3.68 &\textbf{3.22} &3.28 &\textbf{4.08} & \textbf{3.22}&\textbf{3.62}\\
		\hline\hline
	\end{tabular}
\end{table*}

We randomly selected 13 volunteers to score the music generated by using MuseGan, MidiNet\cite{yang2017midinet} and MIDI-Sandwich2 without fusion, do fusion and generate directly and do fusion fusion adaptation, respectively. we asked them to rate these songs from 1 to 5 for 6 indicators. Finally, we calculate an average score for the 6 indicators. It is obvious from Table \ref{tab:aStrangeTable3} that multi-track music generated directly without fusion performs the worst. The music generated directly by MIDI-sandwich2 performs slightly lower than MuseGAN, which shows that the mult-itrack music generated by RNN-based model, MIDI-Sandwich2, is close to the state of art level of multi-track music. The music using MIDI-Sandwich2 to restore and adapt performs the best, meaning that MIDI-Sandwich2 can adapt multi-track music with level beyond the current state of art under the combination of human music and Artificial Intelligence (AI).
The sample of the evaluation result is also provided in \url{https://github.com/LiangHsia/MIDI-S2}.

\section{CONCLUSIONS}
In this paper, we explore the problem that multi-track symbolic music generation cannot be realized using RNN-based generation model. We propose a multi-modal fusion-generation model using a layered VAE architecture. The model is divided into two major modules, BVAE and MFG-VAE. The BVAE, RNN-based VRAE, improves the performance of BMuseGAN by converting the binarization problem into a multi-label classification problem, which solves the problem that the refinement step of the original scheme is difficult to differentiate and the gradient is hard to descent in BMuseGAN. The underlying model BVAE is independent of each other and is used to process different track data. The upper model MFG-VAE adopts the multi-modal fusion-generation VAE architecture. The encoder is used as the multi-modal fusion networks, and the decoder is used as the multi-modal generation networks. Through the fusion/generation of the latent vector generated by the underlying BVAE, the whole model can generates multi-track symbolic music.

Through the model proposed in this paper, we first use the RNN-based model for multi-track music generation, and the effect of generated multi-track music is very close to the current state of the art work, MuseGAN network. This indicates that more RNN network structures, such as LSTM, Gated recurrent units (GRU), and RNN technologies, such as Attention, Seq2seq, etc., can be applied to multi-track music generation by combining with the hierarchical VAE proposed in this paper. At the same time, VAE's unique ability to reconstruct songs makes the songs generated by the MIDI-Sandwich2 retain some features of the original song, and is slightly different from the original song. This semi-automatic adaptation of the music is more beautiful than the generated sample of the MuseGAN. Based on the VAE model, it can also be applied to music style transfer, music mixing, music interaction and other issues. This shows that the VAE model proposed in this paper has certain value in various fields of AI music.

\addtolength{\textheight}{-12cm}   




\bibliographystyle{ieeetran}
\bibliography{ref}

\end{document}